\documentclass{article} 
\usepackage{iclr2027_conference,times}

\usepackage{float}
\usepackage{hyperref}
\usepackage{url}
\usepackage{xcolor}
\usepackage{colortbl}
\usepackage{amsmath}
\usepackage{amssymb}
\usepackage{booktabs}
\usepackage{graphicx}
\usepackage{float}
\usepackage{tabularx}
\usepackage{multirow}

\title{Paired Multimodal Scaling Laws}

\author{Marcus Ma \\
University of Southern California \\
\texttt{mjma@usc.edu} \\
\And
Shrikanth Narayanan \\
University of Southern California \\
}

\iclrfinalcopy
\begin{document}

\maketitle

\begin{abstract}
Existing multimodal scaling laws fit multimodality terms empirically after testing and never vary how much data is multimodally paired at fixed data budgets. We investigate how, under the same total data per modality, changing the number of paired data affects loss curves in multimodal classification tasks. We train models in three different environments and run experiment sweeps varying data sizes and pairing budget. Pairing ratios have a dramatic impact on loss and this impact is directly tied to how much information synergy the task contains. Only paired data is able to reduce synergistic loss, while unpaired data can reduce redundant or unimodal information up until unimodal floors. Unlike traditional scaling laws where loss drops immediately in power law decay, synergy acquisition is gated, requiring a critical threshold of paired data before synergistic loss falls at all. We introduce a new family of multimodal scaling laws where total data-attributable loss is the sum of four individual power laws corresponding to the four different information channels of redundancy, a unique channel per modality, and synergy, and show how this law is both more theoretically sound and empirically valid across our experiments. This law predicts multimodal loss in our experiments more accurately than existing laws (3.2\% error on fit tests versus 10.4\% error for pairing extensions of published laws) and can be used to determine optimal pairing fractions under constrained budgets.
\end{abstract}

\section{Introduction}
\label{sec:intro}

Scaling laws predict how neural networks perform as a function of resources spent on it. This is traditionally represented in terms of model parameters $N$ and training examples $D$ \citep{hestness_2017, kaplan_2020}. \citet{hoffmann_2022} formulate this equation as

\begin{align}
L(N,D) = \underbrace{E}_{\text{irreducible error}} + \underbrace{\frac{A}{N^\alpha}}_{\text{approximation error}} + \underbrace{\frac{B}{D^\beta}}_{\text{estimation error}},
\label{eq:hoffmann-scaling}
\end{align}
where the irreducible error represents the Bayes-optimal theoretical error floor, the approximation error is loss due to finite model size, and the estimation error is loss due to finite data sample.

Multimodal extensions of scaling laws keep the same shape as \citet{hoffmann_2022} but divide $D$ into multiple subcomponents, i.e. by modeling loss as $L(N, D_1, D_2)$. \citet{aghajanyan_2023} decompose each modality into its own unimodal scaling law, then adjust empirically with multimodal interaction constants. Other laws represent loss in terms of $D$ but with additional terms that capture the per-domain fraction of the total data \citep{ye_2024, tao_2026}. These laws represent multimodality by modeling \textit{how much} of each modality is present.

However, a multimodal corpus contains a second quantity that is not modeled in existing scaling laws: \textbf{multimodal pairing}, or the presence of \textbf{both} modalities for a single data point. For an amount of paired data $n_p$, we represent the pairing fraction $p$ as $p = \frac{2n_p}{D_1 + D_2} \in [0, 1].$

$p = 1$ only when every example is a pair and adding unpaired examples of either modality lowers $p$. Existing works have never varied $p$ while holding $D_1$ and $D_2$ fixed, and we hypothesize that their failure to model $p$ explicitly could partially explain why each work arrives at a different scaling law equation fitting their empirical results. When adding extra unpaired data has been tested, it is always in the text modality, and it always has come with the confound of adding extra total data, which means the effect of pairing alone has never been isolated.

We run experiments that vary $p$ across three settings at fixed data budget and find that $p$ has a direct impact on downstream loss that existing laws fail to capture. We formulate new multimodal scaling laws that incorporate this variable based on Partial Information Decomposition \citep{bertschinger_2014, liang_2023} and find our laws to be more theoretically grounded and empirically accurate. In particular, paired data is the only data stream that enables learning of \textbf{synergistic} information, or emergent information that arises only when both modalities are present. We also find unpaired data helps reduce non-synergistic information loss, but only with sufficient paired data alongside. Synergy is also gated: unlike traditional scaling law decay, synergistic loss stays at chance until a certain pair count threshold.

Using these findings, we introduce a new four-channel multimodal scaling law, where data loss is the sum of four independent power law curves:
\begin{align}
L(N, D_1, D_2, n_p) = E + \frac{A}{N^\alpha} + L_R + L_{U_1} + L_{U_2} + L_S
\end{align}
where each $L$ term is fed data only by relevant modalities: redundant $R$ by either modality, unique $U_1$ and $U_2$ by modalities 1 and 2, respectively, and synergistic $S$ only by paired data $n_p$. This family of scaling laws fits our empirical results significantly better than existing multimodal scaling laws and are the only laws capable of modeling the synergistic phenomena we observe. Their form is also more grounded in multimodal interaction theory.

\section{Related Work}
\label{sec:related}

Our work builds on multimodal scaling laws (\S\ref{sec:rel-multimodal}), data-mixture laws (\S\ref{sec:rel-mixture}), and partial information decomposition (\S\ref{sec:rel-pid}), a subset of information theory that motivates our experiments and laws.

\subsection{Multimodal scaling laws}
\label{sec:rel-multimodal}

In unimodal settings (Eq.~\ref{eq:hoffmann-scaling}), while loss is a variable of $N$ and $D$, $A$ and $B$ set the intercept of loss and the exponents $\alpha$ and $\beta$ determine the rate of loss scaling. Exponents are set mainly by the task and the data distribution while architecture mainly moves the starting points $A$ and $B$ \citep{hestness_2017, kaplan_2020}.

\citet{aghajanyan_2023} fits separate scaling curves per modality and bimodal runs to identify a synergy constant $C_{i,j}$ at odds with competition terms in $N$ and $D$. Because synergy is modeled as a constant, a ``competition barrier'' exists at an $N$-$D$ curve where synergy outweighs competition. \citet{shukor_2025} train 457 image-text models across a variety of architectures and find little variation at scale but that early-fusion models perform better at lower parameter counts. Like \citet{aghajanyan_2023}, this alludes to a ``synergy threshold'' at scale. \citet{sun_2024} propose weighting multimodal mixtures by tokenization efficiency and \citet{tao_2026} fit a compute-dependent modality mixture with a learned interaction matrix. Several of these works mix a text-only pool into paired image-text data, which lowers $p$ but also adds data, so the role of $p$ is never isolated.

\subsection{Data-mixture scaling laws}
\label{sec:rel-mixture}

Data mixture laws ask what mixture a corpus, typically text, each domain (web text, books, papers) should be assigned, either fit before training or updated online. \citet{ye_2024} present the law as minimization of individual domain losses and \citet{liu_2025} regress loss on a mixture of proxy models targeting 2\% compute of a target run. \citet{que_2024} investigate the ratio of domains $r$ as a scaling term and \citet{shukor_2025b} find mixture ratios that are fixed regardless of scale perform best. Other approaches operate online; \citet{albalak_2023} presents training as a bandit problem, and \citet{jiang_2024} calculate per-domain power laws during the run. \citet{li_2025} estimate loss bounds per-task via gradient norms. However, \citet{chen_2025} identify no method above consistently beats \textit{uniform sampling} in general. They diagnose that all of the functional forms fail to model interaction terms between mixtures, so no equations generalize.



\subsection{Partial information decomposition}
\label{sec:rel-pid}

Partial Information Decomposition, introduced by \citet{williams_2010}, decomposes the joint mutual information shared by two feature variables and a target variables into $I(X_1, X_2; Y) = R + U_1 + U_2 + S$, where $R$ is redundant information shared between both modalities on $Y$, $U_1$ and $U_2$ are unique information per modality, and $S$ is joint information present only with both modalities. \citet{liang_2023} formulate this problem for multimodal training loss. Given saturated unimodal training losses $L_1$ and $L_2$, multimodal loss $L_{1,2}$, and initial loss $L$, we have:
\begin{align}
    L - L_1 = U_1 + R, \qquad L - L_2 = U_2 + R, \qquad L - L_{1,2} = S +U_1 + U_2 + R
\end{align}

Using linear combinations of these three values, we are able to calculate $R + U_1$, $R+U_2$, and $S-R$, but not $R$ or $S$ in isolation. \citet{bertschinger_2014} and \citet{liang_2023} identify tractable ways to estimate $S$ for joint distributions and multimodal machine learning. We assert that modeling $S$ is the overlooked solution for better multimodal scaling laws. $S$ determines how much co-occurring information is required for a task, which we hypothesize can only be learned during \textit{paired} interaction. Meanwhile, we hypothesize \textit{unpaired} data can help reduce $U_1, U_2,$ and $R$ independently but have \textbf{no impact} on $S$.

\section{Experiments}
\label{sec:experiments}

We create three environments that vary architectures, data, and multimodal interaction. In the first (Digits), we train image and audio encoders to predict digits in a setup where we can adjust the PID information $R, U_1, U_2,$ and $S$. In the second (VQA), we train classification heads on top of pretrained frozen embeddings for visual question answering using VQA \citep{antol_2015}, a multimodal dataset heavily dominated by the text modality \citep{goyal_2017}. In the third (NLVR2), we finetune a pretrained multimodal transformer on NLVR2 \citep{suhr_2019}, another visual question dataset where mutual information is heavily dominated by $S$ and unimodal models stay at chance.

\begin{figure}[t]
\centering
\includegraphics[width=\linewidth]{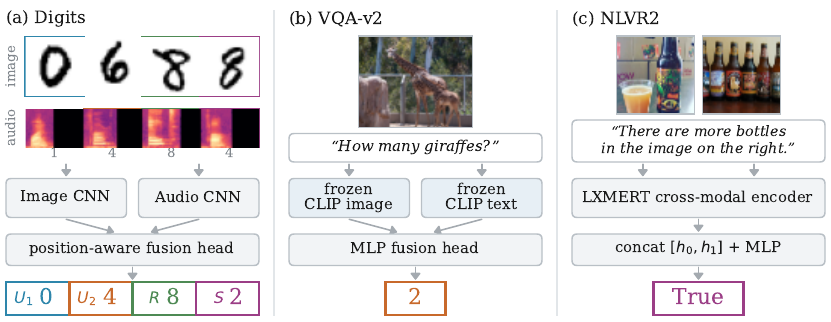}
\caption{Illustration of each environment along with an example and tested architecture.}
\label{fig:env-overview}
\end{figure}

\subsection{Methodology}

We apply the same methodology to each environment. We reserve a fixed, unseen test set for every run. Unpaired examples fill in the missing modality with a learnable null vector and a modality-presence bit. During test time, both modalities are always present and loss is cross-entropy. In Appendix~\ref{app:metrics}, we analyze replacing cross-entropy with accuracy.

\paragraph{Grid sweeps and evaluation.} We run experimental sweeps across data budgets $D$, compute budget $C$, pairing count $n_p$, and multiple seeds per combination. For each experiment, we fix a single architecture and parameter count $N$, as prior scaling law literature has demonstrated that loss due to architecture and parameter counts is independent of loss attributable to data composition \citep{hestness_2017, kaplan_2020, shukor_2025}. For conciseness, we combine the irreducible and approximation errors $E$ and $A/N^\alpha$ from Eq.~\ref{eq:hoffmann-scaling} into a single term $E_N$. In Appendix~\ref{app:architecture}, we repeat experiments for Digits and VQA on different architectures and parameter sizes and again find the same qualitative results.

\paragraph{Counting paired and unpaired data.} We determine total data via $D = D_1 + D_2$, i.e. the sum of unique examples per modality in the dataset. This means as $p\rightarrow 1$, the number of unique rows decreases. We do \textbf{not} split a pair into two unpaired data: each row is a unique label. As this manner of counting is more favorable to unpaired data, any improvement from $p=0$ to $p=1$ is tied directly to pairing. We fix equal compute via training step count, so a model trained at $D_{p=1}$ has twice as many training epochs as $D_{p=0}$ but the same gradient steps. In Appendix~\ref{app:counting}, we consider other ways of counting paired and unpaired data.

\subsection{The Three Environments}

Figure~\ref{fig:env-overview} and Table~\ref{tab:exp-primitives} illustrate each environment and architecture. We describe each experimental setup below with implementation details in Appendix~\ref{app:experiment-implementation}.

\paragraph{Digits.} We construct series of four concatenated digits over MNIST images \citep{lecun_1998} and AudioMNIST spectrograms \citep{becker_2024} The label is a four-digit number corresponding directly to $U_1$, $U_2$, $R$, and $S$ for each position: the first and second digits ($U_1$ and $U_2$) are the first and second digit shown in image and audio, the third digit ($R$) is the image and audio's third digit value, and the fourth digit ($S$) are the fourth input digits' sum $\mod 10$, so knowing only one modality is uninformative. We add symmetric label noise $\epsilon = 0.1$ so $U_1 = U_2 = R = S = \ln 10 - H_{\epsilon=0.1} \approx 1.758$. Loss is averaged over each label digit. The model architecture is two CNN encoders that pool to one value per position then a fused position-aware head.

\begin{table*}[t]
\centering
\scriptsize
\begin{tabular}{@{}lcccccc@{}}
\toprule
& \multicolumn{2}{c}{Digits} & \multicolumn{2}{c}{VQA-v2} & \multicolumn{2}{c}{NLVR2} \\
\cmidrule(lr){2-3}\cmidrule(lr){4-5}\cmidrule(lr){6-7}
& image & audio & image & question & photos & sentence \\
\midrule
\multicolumn{7}{@{}l}{\emph{Experimental details}} \\
architecture & \multicolumn{2}{c}{CNN + 4-way head ($476$k)} & \multicolumn{2}{c}{Frozen CLIP + head ($5.3$M)} & \multicolumn{2}{c}{Finetuned LXMERT ($210$M)} \\
chance loss $L_0$ & \multicolumn{2}{c}{$\ln 10 \approx 2.303$} & \multicolumn{2}{c}{$\ln 3129 \approx 8.048$} & \multicolumn{2}{c}{$\ln 2 \approx 0.693$} \\
$D$
  & \multicolumn{2}{c}{$8$k, $16$k, $32$k, $64$k, $128$k, $256$k}
  & \multicolumn{2}{c}{$0$, $4$k \dots\, $256$k}
  & \multicolumn{2}{c}{$4$k, $8$k, $16$k, $32$k, $64$k} \\
$C$ ($k(D_1+D_2)$ steps)
  & \multicolumn{2}{c}{$k \in \{10,20,40\}$}
  & \multicolumn{2}{c}{$k \in \mathbb{Z_+} \le 80$}
  & \multicolumn{2}{c}{$k \in \{1,2,3\}$} \\
runs & \multicolumn{2}{c}{$408$ ($3$ seeds)} & \multicolumn{2}{c}{$3{,}660$ ($5$ seeds)} & \multicolumn{2}{c}{$1{,}010$ ($5$ seeds)} \\
\midrule
\multicolumn{7}{@{}l}{\emph{Unimodal data curve fitting $L(D) = E_N + B \cdot D^{-\beta}$ at the above $N$}} \\
$E_N$ & $0.573$ & $0.587$ & $4.169$ & $2.683$ & $\approx\ln 2$ & $\approx\ln 2$ \\
$B$ & $69.7$ & $119$ & $2.17$ & $59.8$ & $>0$ & $>0$ \\
$\beta$ & $0.684$ & $0.732$ & $0.095$ & $0.411$ & $\approx \infty$ & $\approx \infty$ \\
\midrule
\multicolumn{7}{@{}l}{\emph{Partial information decomposition (nats)}} \\
$R$   & \multicolumn{2}{c}{$1.758$} & \multicolumn{2}{c}{$0.531$} & \multicolumn{2}{c}{$0$} \\
$U_1$ & \multicolumn{2}{c}{$1.758$} & \multicolumn{2}{c}{$0.127$} & \multicolumn{2}{c}{$0$} \\
$U_2$ & \multicolumn{2}{c}{$1.758$} & \multicolumn{2}{c}{$1.614$} & \multicolumn{2}{c}{$0$} \\
$S$   & \multicolumn{2}{c}{$1.758$} & \multicolumn{2}{c}{$0.144$} & \multicolumn{2}{c}{$0.532$} \\
\bottomrule
\end{tabular}
\caption{Experimental details, unimodal data curve variables, and PID terms for each environment. Compute $C$ represents number of training steps multiplied by batch size.}
\label{tab:exp-primitives}
\end{table*}

\paragraph{VQA.} We utilize VQA-v2 train, a collection of images and questions with labels in a 3,129 label vocabulary. We freeze CLIP ViT-B/32 \citep{radford2021clip} image and caption embeddings. We then train a fusion head with unpaired images and captions accompanied with a learned null vector. As initially reported by \citet{goyal_2017}, the VQA dataset is heavily dominated by the question text alone.

\paragraph{NLVR2.} NLVR2 is another image-question dataset engineered so multimodality is required. Each example is two photographs plus a sentence that asks a comparative question. All questions are binary with even splits between True/False, so individual unimodal accuracy hovers around 50\%. We finetune an existing model, LXMERT \cite{tan_2019}, which has achieved 74.45\% on NLVR2, using the published model code\footnote{\url{https://github.com/airsplay/lxmert}}.

\paragraph{Unimodal loss curves and PID.}
\label{sec:exp-pid}

We run large-scale unimodal experiment runs until loss saturation and determine unimodal loss curves for each modality with Hoffmann-style variable estimates in Table~\ref{tab:exp-primitives}. We use these curves and task setup to estimate PID variables $R$, $U_1$, $U_2$, and $S$. For Digits, this is analytically know to be $\approx 1.758$ nats. For VQA, the mutual information terms $U_1$, $U_2$, and $R$ are estimated as linear combinations between the soft-label entropy $H(y)=4.828$ nats minus the cross-entropy of saturated unimodal runs and a multimodal run, following \citet{bertschinger_2014}. Shuffling questions within majority-answer class raises paired loss from $2.411$ to $2.556$ which estimates $S$ as $0.144$. NLVR2's unimodal probes always stay at chance, which we use to infer that 100\% of reducible loss comes from synergy. We set $S=0.532$ as the lower bound of mutual information based on the published human accuracy of $96.2\%$ on the development split \citep{suhr_2019}.

\section{How Does Pairing Affect Loss?}
\label{sec:paired-curves}

\begin{figure}[t]
\centering
\includegraphics[width=\linewidth]{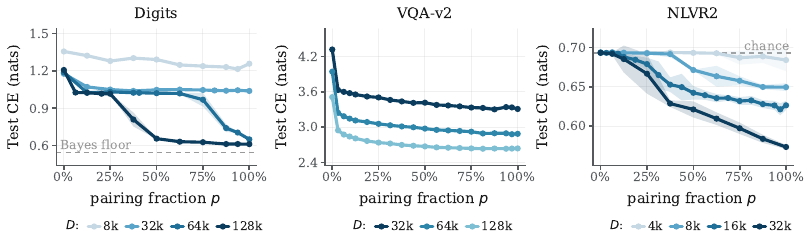}
\caption{Test cross-entropy loss versus pairing fraction $p$ at fixed per-modality data budgets $D_1=D_2$ and fixed compute ($k \cdot D$ gradient steps; $k=20$ for Digits and VQA, $k=3$ for NLVR2). $\pm 1$ std shown as shaded bands. For NLVR2, we filter out ``failed-to-start'' seeds, as discussed in \S\ref{sec:synergy_gates}.}
\label{fig:pair-curves}
\end{figure}

Paired data, and only paired data, is capable of reducing synergistic loss. Meanwhile, information conveyable in single modalities is learnable from unpaired data, but models require paired data to transfer unpaired learnings to a paired setting. In Figure~\ref{fig:pair-curves}, we keep total data per-modality fixed and equal ($D_1=D_2$) and vary percentage of paired data. Increased pairing decreases loss in all three environments, and we hypothesize the relative percentage of reducible loss depends on task synergy after an initial one-time loss drop following $p=0$. We identify this initial drop as pairs unlocking unpaired data: unpaired data is only useful in multimodal settings after learning paired interaction.

In the Digits environment, for high $D$, we notice a \textbf{plateau then second descent}. This behavior reflects the model learning $R$, $U_1$, and $U_2$ labels very quickly and taking a longer time to learn $S$; a similar flatline then drop arises in NLVR2. Additionally, for NLVR2, some seeds never leave chance, with the likelihood of a ``failed-to-start'' seed increasing at lower $p$ and lower $D$. We discuss these phenomena further in \S\ref{sec:synergy_gates}. In VQA, where synergy accounts for 6\% of loss, there is a large loss dropoff from $p=0\%$ to $p=5\%$ but after a much more modest reduction in loss as $p \rightarrow 1$. At compute higher than $k=20$ for VQA, optimal $p$ lies interior of $p=1$, which we note appears to be an artifact of overfitting because total examples halve as $p\rightarrow 1$. We conclude at fixed $D$, optimal compute decreases with increased $p$ because repeat gradient steps per datum increase. 

\subsection{Controlling for Modality Presence Matching}
\label{subsec:presence-matching}

\begin{figure}[t]
\centering
\includegraphics[width=\linewidth]{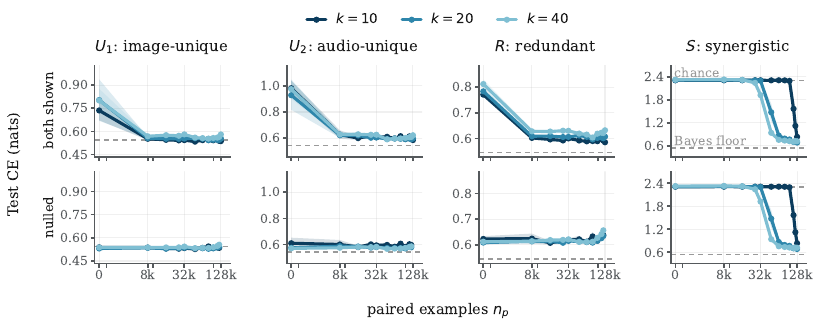}
\caption{Individual channel losses at $D_1=D_2=128k$ depending on which modalities are shown at test time. \textbf{Top:} loss when every test example shows both modalities. \textbf{Bottom} loss when irrelevant modalities are nulled for that given channel. For $U_1$, audio is nulled, for $U_2$, image is nulled, for $R$, both are nulled then averaged, and for $S$, no change.}
\label{fig:pair-slots}
\end{figure}

One confound from above is that in our setup, each example in the test set always contains both modalities, so improvement in $p$ could simply be train-test alignment. This explains the significant jump in performance following $p=0$ for Digits and VQA, because at $p=0$ there are zero paired examples to learn from. In order to test this, we re-run Digits runs where test examples are also unpaired, i.e. when either modality is nulled out. In Figure~\ref{fig:pair-slots}, we plot loss of each information channel twice, on top when both modalities are present and on the bottom when irrelevant modalities are nulled: for $U_1$, we null audio, for $U_2$, the image, and for $R$, both individually with the two losses averaged. When controlling for presence matching in this fashion, increasing pairing fraction has \textbf{no impact} on $U_1$ and $U_2$, and \textbf{slightly increases the loss} of $R$. We hypothesize that as $p\rightarrow 1$, the redundant prediction learns to rely on both channels simultaneously, so zeroing out a single channel actually hurts performance rather than having no effect like for $U_1$ and $U_2$. At low $p$, showing both modalities \textbf{increases loss} for single-modality channels ($U_1$, $U_2$, $R$) compared to just showing one modality because the model has not been exposed to enough paired examples in training to learn the other modality is irrelevant. After around $8$k pairs, though, the top and bottom readouts converge to the same loss. We conclude, then, when controlling for train-test modality presence matching, paired data only helps reduce synergistic loss, but in general, pairing is required for unpaired training data to be useful during paired deployment.

\subsection{When Does Extra Unpaired Data Help?}

\begin{figure}[t]
\centering
\includegraphics[width=\linewidth]{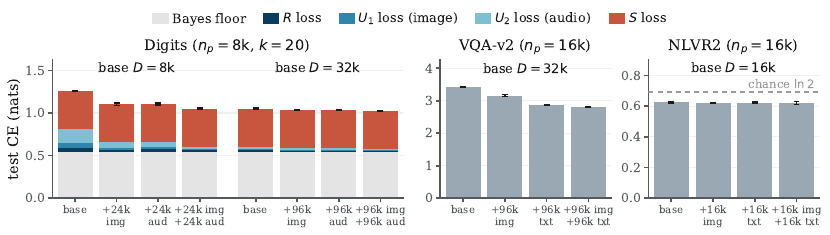}
\caption{Test losses across the three environments at fixed $n_p$ and $k$ epochs, with additional unpaired data added. Digits loss is decomposed into individual information channel losses and the Bayes floor.}
\label{fig:unpaired-nats}
\end{figure}

Figure~\ref{fig:pair-curves} holds a total data budget fixed. In Figure~\ref{fig:unpaired-nats}, we fix paired data and add extra unpaired examples. For Digits, we subdivide loss into the Bayes floor and each individual information channel's loss. Unpaired data reduces redundant and modality-unique loss but never synergistic loss. Additional unpaired data has the highest impact at lower overall $D$, with 24k more images or audio clips reducing loss at $D=8$k from 1.26 to 1.10 nats; however, once the unimodal floors are more saturated at $D=32$k, adding the same ratio barely affects loss. In both settings, synergistic loss is not reduced. This is consistent with NLVR2 seeing no improvement with additional unpaired data, since NVLR2 is dominated by $S$. Meanwhile, for VQA, adding unpaired text examples reduces loss more than images, which is consistent with relative PID weights for text and images. How much unpaired data helps also depends on the number of pairs; when $n_p=128$, $124$k unpaired questions reduce loss by $0.92$ nats, but at $n_p=4$k, it reduces loss by $1.20$ nats. Pairs and unpaired data are therefore complements initially: adding some of one pool raises the value of the other. We use these results to conclude that \textbf{unpaired data does help reduce loss given sufficient pairs, but only to reduce $R$, $U_1$, and $U_2$ losses.}

\subsection{Synergy Acquisition is Gated}
\label{sec:synergy_gates}

Both the Digits $S$ channel and the NLVR2 loss curves display atypical behavior at onset: they sit at chance initially then drop after a certain threshold of $n_p$. Meanwhile, VQA, whose synergistic loss share is small, and the $R$, $U_1$, and $U_2$ Digit channels, exhibit no such behavior. Synergy acquisition is gated on two factors: a minimum of paired examples $n_{\text{floor}}$ and a minimum of paired exposures (including repeats) $r$. We propose the synergy gate $n^*$ is the number of paired data required before synergistic loss drops, $n^* = \max(n_{\text{floor}}, r/k)$.

\begin{figure}[t]
\centering
\includegraphics[width=\linewidth]{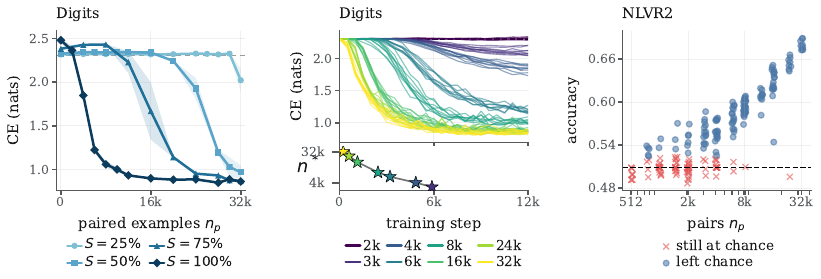}
\caption{Evidence of a synergy gate. \textbf{(a)} Digits loss of the $S$ channel as a factor of $n_p$ and how much relative weight the $S$ channel contributes to total loss at $D=64k$ and $k=40$. \textbf{(b)} \textbf{Top:} Digits loss of the $S$ channel as a factor of $n_p$ and training step (batch size$=128$). \textbf{Bottom:} median training step and $n^*$ per data budget when $S$ loss leaves chance. $n^*$ increases with higher data budgets because total repetitions lower. \textbf{(c)} NLVR2's ``failed-to-start'' seeds (red X's) versus seeds that leave chance accuracy (blue dots), plotted against $n_p$.}
\label{fig:synergy-gate}
\end{figure}

For Digits, $n_\text{floor} \approx 2900$ and $r \approx 40k$ for an interleaved data mixture of paired and unpaired data. We find data presentation order matters. When paired data is entirely frontloaded, synergy is acquired after the paired data but forgotten by the end of training, similar to catastrophic interference \citep{mccloskey_1989}. When paired data is entirely backloaded, synergy is acquired, but at a worse rate: synergistic loss drops $0.60$ nats during interleaved but only $0.15$ nats during unpaired-then-paired training. In Figure~\ref{fig:synergy-gate}, we display several properties of this gate. $n^*$ is influenced by how much of the total loss is composed of synergy. As total loss share of $S$ moves from $25\%$ to $100\%$, $n^*$ decreases: when $S$ loss constitutes half of total loss, $n^*$ is halved to 20k, and when $S$ loss makes up $75\%$, $n^*$ decreases to 12k. Synergy gating occurs NLVR2, but unlike Digits, it is not a hard threshold gate but a random variable of $n^*$. NLVR2 random chance escape is best modeled as $\log_2 n^* \sim \mathcal{N}(11.32,\ 1.36^{2})$,
which models the actual escape distribution with median $n^* = 2552$ and a $\chi^2$ fit of $p=0.81$.

\section{Paired Scaling Laws}
\label{sec:theories}
\suppressfloats[t]

We now attempt to form a generalizable multimodal scaling law with pairing as an input parameter, $L(N, D_1, D_2, n_p)$. We propose several extensions of existing scaling laws by adding $p$ as a variable where at $p=1$ they simplify to the original equation. We then introduce a new loss family, where data loss is summed as four separate loss curves based on $R$, $U_1$, $U_2$, and $S$, and permute four law variations of this family. These new laws consistently outperform pairing-variable extensions of published laws, even at $p=1$. We provide theoretical backing for this (namely, that existing laws cannot model synergy) and show how these laws can find optimal pairing ratios under a pairing budget where pairs are more expensive than unpaired data.

\begin{table*}[t]
\centering
\scriptsize
\resizebox{\textwidth}{!}{%
\begin{tabular}{@{}l c c c l l@{}}
\toprule
Law & $k$ & $\Gamma$ & $\partial\Gamma/\partial D \neq 0$ & $L(N,D_1,D_2,n_p)$ & at $p=1$ \\
\midrule
Hoffmann Unimodal & 4 & $\le 0$ & $\checkmark$ & $E_N + B\,g(D_1{+}D_2)$ & $E_N + B\,g(2D)$ \\
Hoffmann on unique $D$ & 4 & $\le 0$ & $\checkmark$ & $E_N + B\,g(D_{\mathrm{tot}})$ & $E_N + B\,g(D)$ \\
Hoffmann on $D_1$ and $D_2$ & 7 & $=0$ & $\times$ & $E_N + B_1 g_1(D_1) + B_2 g_2(D_2)$ & $E_N + B_1 g_1(D) + B_2 g_2(D)$ \\
Aghajanyan $+Cp$ & 8 & $=0$ & $\times$ & $E_N + B_1 g_1(D_1) + B_2 g_2(D_2) + C p$ & $E_N + B_1 g_1(D) + B_2 g_2(D) + C$ \\
Shukor mixture & 6 & $<0$ & $\checkmark$ & $E_N + (C_1 p + C_2(1{-}p))^{-1} + B\,g(D_{\mathrm{tot}})$ & $E_N + C_1^{-1} + B\,g(D)$ \\
Ye mixture & 7 & any & $\checkmark$ & $E_N + C\exp(\alpha_2 n_2/D_{\mathrm{tot}} + \alpha_p n_p/D_{\mathrm{tot}}) + B\,g(D_{\mathrm{tot}})$ & $E_N + C\,e^{\alpha_p} + B\,g(D)$ \\
\midrule
\emph{Four-channel family} & $k$ / $\Delta k$ & & & \emph{base law and switch substitutions} & \emph{what it models} \\
Base & 14 & $\le 0$ & \multirow{5}{*}{$\checkmark$} & $E_N + \sum_a B_a\, g_a(D_a)$ & one curve per PID channel \\
Shared & $-6$ & no change &  & $g_a \to g$ & one $(\delta,\beta)$ for all channels \\
Gate & $+0$ & any &  & $B_S\, g_S(n_p) \to B_S/(1{+}(n_p/n_\star)^\gamma)$ & synergy is gated \\
PID & $-3$ & no change &  & $B_a \to s_0 W_a$ & amplitudes from measured PID \\
Effective Count & $+2$ & any &  & $D_i \to D_i - (1{-}\lambda_0)\, n_i\, \frac{n_\lambda}{n_p+n_\lambda}$ & pairs unlock unpaired data \\
\bottomrule
\end{tabular}}%
\caption{Extensions of existing scaling laws (Top) and the introduced four-channel law and variants (Bottom). Existing laws reduce to their original equations at $p=1$. Each of the variations of the four-channel law can be applied independently, resulting in 16 four-channel law combinations. $k$ is the number of fitted parameters for the corresponding law, $D_\mathrm{tot}=D_1+D_2-n_p$, $n_i=D_i-n_p$, and $W_a$ are measured PID weights. $\Gamma = -\partial^2 L/\partial D_1 \partial D_2$ is taken at fixed $p$; for the four-channel family its sign is that of the base law unless a switch changes it. $\Gamma$ and $\partial \Gamma / \partial D$ are discussed in \S\ref{subsec:law-results}.}
\label{tab:law-candidates}
\end{table*}

\subsection{Candidate Laws}
\label{sec:law-candidates}

\paragraph{Extending existing laws.} We provide six candidates from existing laws (Table~\ref{tab:law-candidates}, Top) and extend their forms to model pairing. The first three are traditional \citet{hoffmann_2022}-style equations with different approaches of counting data: as the total data exposures per modality, the number of unique data rows, and two independent data curves per modality. We then extend an existing multimodal scaling law and two data mixture laws from the literature: \citet{aghajanyan_2023}'s law that adds a constant ``synergy'' factor, \citet{shukor_2025b}'s law that provides a ``usefulness'' term to each mixture (paired and unpaired), and \citet{ye_2024}'s exponential of the mixture proportions.

\paragraph{The four-channel law.} Building on our analysis that $R, U_1, U_2,$ and $S$ explains much of the variation between environments, we present an alternative equation family, that \textbf{total loss can be broken down as the sum of independent power law curves for each information channel}:

\begin{align}
    L(N, D_1, D_2, n_p) = E_N + L_R + L_{U_1} + L_{U_2} + L_S
\end{align}

Each individual $L$ term is a data power curve where the total amount of ``data'' for that term corresponds to the number of examples relevant to its channel. $R$ is fed by $D_1 + \kappa D_2$, where $\kappa$ is the learned relative usefulness per modality, and $U_1$, $U_2$ and $S$ are fed by $D_1$, $D_2$, and $n_p$. As we fix architecture and model size in our experiments, $E_N=E+A/N^\alpha$ is a constant. We vary four factors across this equation family for a total of sixteen variations:
\begin{enumerate}
    \item Shared $\delta$ and $\beta$: whether the data curve intercepts and exponents are shared across the four losses or independent per curve
    \item Synergy offset: whether the synergy gating effect is modeled in $L_S$ via a soft gate $1/(1+(n_p/n^*)^\gamma)$, based on \S\ref{sec:synergy_gates}
    \item PID-pinned: whether loss terms' relative weights are set by the environment's PID ratios
    \item Effective counts: whether unpaired example count is discounted until a critical threshold of paired data, based on the ``pairs unlock unpaired data'' finding from \S\ref{subsec:presence-matching}
\end{enumerate}

We replace $B/D^\beta$ with $(L_0-E_N)\bigl(\delta/(D+\delta)\bigr)^\beta$, which converge at $D >> \delta$, for better-behaved curves as $D\rightarrow 0$ and let $g(D) = (\delta/(D+\delta))^\beta$ for conciseness. Full derivations in Appendix~\ref{app:paired-laws}.

\begin{table*}[t]
\centering
\scriptsize
\definecolor{lawred2}{RGB}{198,96,96}
\definecolor{lawred1}{RGB}{246,210,210}
\definecolor{lawgreen1}{RGB}{208,234,212}
\definecolor{lawgreen2}{RGB}{96,168,112}
{
\setlength{\aboverulesep}{0pt}
\setlength{\belowrulesep}{0pt}
\renewcommand{\arraystretch}{1.0}
\resizebox{\textwidth}{!}{%
\begin{tabular}{@{}c *{4}{c} c *{12}{wc{0.82cm}}@{}}
\toprule
& \multicolumn{4}{c}{} & & \multicolumn{3}{c}{Digits} & \multicolumn{3}{c}{VQA} & \multicolumn{3}{c}{NLVR2} & \multicolumn{3}{c}{Overall} \\
\cmidrule(lr){7-9}\cmidrule(lr){10-12}\cmidrule(lr){13-15}\cmidrule(lr){16-18}
& \multicolumn{4}{c}{Law} & $k$ & $D{\uparrow}$ & $p{=}1$ & $R^{2}$ & $D{\uparrow}$ & $p{=}1$ & $R^{2}$ & $D{\uparrow}$ & $p{=}1$ & $R^{2}$ & $D{\uparrow}$ & $p{=}1$ & $R^{2}$ \\
\midrule
 & \multicolumn{4}{l}{Hoffmann Unimodal} & 4 & \cellcolor{lawred2}13.9\% & \cellcolor{lawred2}30.4\% & 0.00 & \cellcolor{lawred2}8.2\% & \cellcolor{lawred2}8.1\% & \cellcolor{lawred2}0.00 & \cellcolor{lawred2}1.8\% & \cellcolor{lawred2}6.6\% & \cellcolor{lawred2}0.00 & \cellcolor{lawred2}8.0\% & \cellcolor{lawred2}15.0\% & \cellcolor{lawred2}0.00 \\
 & \multicolumn{4}{l}{Hoffmann on unique $D$} & 4 & \cellcolor{lawred2}14.9\% & \cellcolor{lawred2}37.1\% & \cellcolor{lawred1}-0.12 & \cellcolor{lawred2}10.5\% & \cellcolor{lawred2}13.1\% & \cellcolor{lawred2}-1.26 & \cellcolor{lawred2}1.7\% & \cellcolor{lawred2}6.9\% & \cellcolor{lawred2}-0.45 & \cellcolor{lawred2}9.0\% & \cellcolor{lawred2}19.0\% & \cellcolor{lawred2}-0.61 \\
 & \multicolumn{4}{l}{Hoffmann on $D_1$ and $D_2$} & 7 & \cellcolor{lawred1}11.5\% & \cellcolor{lawred1}27.3\% & 0.00 & \cellcolor{lawgreen1}2.7\% & \cellcolor{lawred1}6.7\% & \cellcolor{lawred2}0.00 & \cellcolor{lawred2}1.8\% & \cellcolor{lawred2}6.6\% & \cellcolor{lawred2}0.00 & 5.3\% & \cellcolor{lawred1}13.5\% & \cellcolor{lawred2}0.00 \\
 & \multicolumn{4}{l}{Aghajanyan $+Cp$} & 8 & \cellcolor{lawred1}11.5\% & 20.4\% & 0.39 & \cellcolor{lawgreen1}2.3\% & 5.0\% & 0.40 & \cellcolor{lawred2}1.5\% & \cellcolor{lawred2}6.0\% & \cellcolor{lawred1}0.22 & 5.1\% & 10.4\% & 0.34 \\
 & \multicolumn{4}{l}{Shukor mixture} & 6 & 9.6\% & \cellcolor{lawred1}29.1\% & 0.46 & \cellcolor{lawred2}8.2\% & \cellcolor{lawred2}9.4\% & \cellcolor{lawred2}0.02 & \cellcolor{lawred1}1.4\% & \cellcolor{lawred2}6.2\% & \cellcolor{lawred2}-0.09 & \cellcolor{lawred2}6.4\% & \cellcolor{lawred2}14.9\% & \cellcolor{lawred1}0.13 \\
 & \multicolumn{4}{l}{Ye mixture} & 7 & 9.9\% & \cellcolor{lawred2}31.8\% & 0.46 & \cellcolor{lawred1}5.3\% & \cellcolor{lawred2}8.1\% & \cellcolor{lawred1}0.06 & 1.3\% & \cellcolor{lawred1}5.7\% & \cellcolor{lawred2}0.11 & 5.5\% & \cellcolor{lawred2}15.2\% & \cellcolor{lawred1}0.21 \\
\midrule
\multirow{17}{*}{\rotatebox[origin=c]{90}{\small Four-channel}} & Shared & Gate & PID & Eff. & \multicolumn{13}{c}{} \\
\cmidrule(lr){2-2}\cmidrule(lr){3-3}\cmidrule(lr){4-4}\cmidrule(lr){5-5}
 & \textcolor{lawred2}{$\times$} & \textcolor{lawred2}{$\times$} & \textcolor{lawred2}{$\times$} & \textcolor{lawred2}{$\times$} & 14 & 9.6\% & 21.4\% & 0.63 & 3.0\% & 3.4\% & \cellcolor{lawgreen1}0.80 & \cellcolor{lawgreen1}\textbf{0.8\%} & 4.7\% & 0.61 & 4.5\% & 9.9\% & \cellcolor{lawgreen1}0.68 \\
 & \textcolor{lawred2}{$\times$} & \textcolor{lawred2}{$\times$} & \textcolor{lawred2}{$\times$} & \textcolor{lawgreen2}{$\checkmark$} & 16 & 9.5\% & \cellcolor{lawgreen2}9.9\% & \cellcolor{lawgreen1}0.75 & 3.2\% & \cellcolor{lawgreen2}\textbf{1.8\%} & \cellcolor{lawgreen1}\textbf{0.86} & \cellcolor{lawgreen1}\textbf{0.8\%} & 4.8\% & 0.61 & 4.5\% & \cellcolor{lawgreen2}5.5\% & \cellcolor{lawgreen1}0.74 \\
 & \textcolor{lawred2}{$\times$} & \textcolor{lawgreen2}{$\checkmark$} & \textcolor{lawred2}{$\times$} & \textcolor{lawred2}{$\times$} & 14 & 9.5\% & 21.7\% & \cellcolor{lawred2}-1.27 & 2.9\% & \cellcolor{lawgreen1}3.0\% & \cellcolor{lawgreen1}0.81 & 1.1\% & \cellcolor{lawgreen1}2.8\% & \cellcolor{lawgreen1}\textbf{0.81} & 4.5\% & 9.2\% & \cellcolor{lawred1}0.12 \\
 & \textcolor{lawred2}{$\times$} & \textcolor{lawgreen2}{$\checkmark$} & \textcolor{lawred2}{$\times$} & \textcolor{lawgreen2}{$\checkmark$} & 16 & 9.5\% & \cellcolor{lawgreen2}\textbf{3.3\%} & \cellcolor{lawgreen1}\textbf{0.95} & 3.8\% & 3.5\% & \cellcolor{lawgreen1}\textbf{0.86} & 1.1\% & \cellcolor{lawgreen1}3.0\% & \cellcolor{lawgreen1}\textbf{0.81} & 4.8\% & \cellcolor{lawgreen2}3.3\% & \cellcolor{lawgreen2}\textbf{0.87} \\
 & \textcolor{lawgreen2}{$\checkmark$} & \textcolor{lawred2}{$\times$} & \textcolor{lawred2}{$\times$} & \textcolor{lawred2}{$\times$} & 8 & \cellcolor{lawgreen1}9.2\% & 23.9\% & 0.66 & 2.9\% & \cellcolor{lawgreen1}2.1\% & \cellcolor{lawgreen1}0.80 & \cellcolor{lawgreen1}\textbf{0.8\%} & 4.7\% & 0.61 & \cellcolor{lawgreen1}4.3\% & 10.2\% & \cellcolor{lawgreen1}0.69 \\
 & \textcolor{lawgreen2}{$\checkmark$} & \textcolor{lawred2}{$\times$} & \textcolor{lawred2}{$\times$} & \textcolor{lawgreen2}{$\checkmark$} & 10 & \cellcolor{lawgreen1}9.2\% & 23.9\% & \cellcolor{lawgreen1}0.74 & 2.9\% & \cellcolor{lawgreen1}2.4\% & \cellcolor{lawgreen1}0.84 & \cellcolor{lawgreen1}\textbf{0.8\%} & 4.7\% & 0.61 & \cellcolor{lawgreen1}4.3\% & 10.3\% & \cellcolor{lawgreen1}0.73 \\
 & \textcolor{lawgreen2}{$\checkmark$} & \textcolor{lawgreen2}{$\checkmark$} & \textcolor{lawred2}{$\times$} & \textcolor{lawred2}{$\times$} & 10 & \cellcolor{lawgreen1}9.2\% & 21.6\% & \cellcolor{lawred2}-1.56 & 2.9\% & \cellcolor{lawgreen1}2.8\% & \cellcolor{lawgreen1}0.81 & 1.1\% & \cellcolor{lawgreen2}2.7\% & \cellcolor{lawgreen1}\textbf{0.81} & \cellcolor{lawgreen1}4.4\% & 9.0\% & \cellcolor{lawred2}0.02 \\
 & \textcolor{lawgreen2}{$\checkmark$} & \textcolor{lawgreen2}{$\checkmark$} & \textcolor{lawred2}{$\times$} & \textcolor{lawgreen2}{$\checkmark$} & 12 & \cellcolor{lawgreen1}9.3\% & \cellcolor{lawgreen2}\textbf{3.3\%} & \cellcolor{lawgreen1}\textbf{0.95} & 3.2\% & 3.5\% & \cellcolor{lawgreen1}0.85 & 1.1\% & \cellcolor{lawgreen1}2.8\% & \cellcolor{lawgreen1}\textbf{0.81} & 4.5\% & \cellcolor{lawgreen2}\textbf{3.2\%} & \cellcolor{lawgreen2}\textbf{0.87} \\
 & \textcolor{lawred2}{$\times$} & \textcolor{lawred2}{$\times$} & \textcolor{lawgreen2}{$\checkmark$} & \textcolor{lawred2}{$\times$} & 11 & \cellcolor{lawgreen2}\textbf{8.6\%} & 21.5\% & 0.63 & \cellcolor{lawgreen1}2.3\% & \cellcolor{lawgreen1}2.7\% & \cellcolor{lawgreen1}0.81 & \cellcolor{lawgreen1}\textbf{0.8\%} & 4.6\% & 0.61 & \cellcolor{lawgreen1}\textbf{3.9\%} & 9.6\% & \cellcolor{lawgreen1}0.68 \\
 & \textcolor{lawred2}{$\times$} & \textcolor{lawred2}{$\times$} & \textcolor{lawgreen2}{$\checkmark$} & \textcolor{lawgreen2}{$\checkmark$} & 13 & \cellcolor{lawgreen1}9.0\% & 21.5\% & \cellcolor{lawgreen1}0.74 & \cellcolor{lawgreen1}\textbf{2.1\%} & \cellcolor{lawgreen1}2.2\% & \cellcolor{lawgreen1}\textbf{0.86} & \cellcolor{lawgreen1}\textbf{0.8\%} & 4.6\% & 0.61 & \cellcolor{lawgreen1}4.0\% & 9.4\% & \cellcolor{lawgreen1}0.74 \\
 & \textcolor{lawred2}{$\times$} & \textcolor{lawgreen2}{$\checkmark$} & \textcolor{lawgreen2}{$\checkmark$} & \textcolor{lawred2}{$\times$} & 11 & 9.8\% & 21.4\% & \cellcolor{lawred2}-1.10 & \cellcolor{lawgreen1}2.4\% & \cellcolor{lawgreen1}3.1\% & \cellcolor{lawgreen1}0.79 & 1.1\% & \cellcolor{lawgreen2}\textbf{2.1\%} & \cellcolor{lawgreen1}\textbf{0.81} & 4.4\% & 8.9\% & \cellcolor{lawred1}0.17 \\
 & \textcolor{lawred2}{$\times$} & \textcolor{lawgreen2}{$\checkmark$} & \textcolor{lawgreen2}{$\checkmark$} & \textcolor{lawgreen2}{$\checkmark$} & 13 & 9.8\% & \cellcolor{lawgreen2}8.5\% & \cellcolor{lawgreen1}\textbf{0.95} & \cellcolor{lawgreen1}2.6\% & 3.5\% & \cellcolor{lawgreen1}\textbf{0.86} & 1.1\% & \cellcolor{lawgreen2}\textbf{2.1\%} & \cellcolor{lawgreen1}\textbf{0.81} & 4.5\% & \cellcolor{lawgreen2}4.7\% & \cellcolor{lawgreen2}\textbf{0.87} \\
 & \textcolor{lawgreen2}{$\checkmark$} & \textcolor{lawred2}{$\times$} & \textcolor{lawgreen2}{$\checkmark$} & \textcolor{lawred2}{$\times$} & 5 & 10.4\% & 22.5\% & 0.49 & 3.2\% & \cellcolor{lawred1}6.2\% & \cellcolor{lawred1}0.22 & \cellcolor{lawgreen1}\textbf{0.8\%} & 4.6\% & 0.61 & 4.8\% & 11.1\% & 0.44 \\
 & \textcolor{lawgreen2}{$\checkmark$} & \textcolor{lawred2}{$\times$} & \textcolor{lawgreen2}{$\checkmark$} & \textcolor{lawgreen2}{$\checkmark$} & 7 & 10.6\% & 22.5\% & 0.54 & 2.8\% & 4.7\% & 0.61 & \cellcolor{lawgreen1}\textbf{0.8\%} & 4.6\% & 0.61 & 4.7\% & 10.6\% & 0.59 \\
 & \textcolor{lawgreen2}{$\checkmark$} & \textcolor{lawgreen2}{$\checkmark$} & \textcolor{lawgreen2}{$\checkmark$} & \textcolor{lawred2}{$\times$} & 7 & 9.8\% & 23.3\% & 0.63 & 3.0\% & \cellcolor{lawgreen1}3.2\% & \cellcolor{lawgreen1}0.79 & 1.1\% & \cellcolor{lawgreen2}\textbf{2.1\%} & \cellcolor{lawgreen1}\textbf{0.81} & 4.6\% & 9.5\% & \cellcolor{lawgreen1}0.74 \\
 & \textcolor{lawgreen2}{$\checkmark$} & \textcolor{lawgreen2}{$\checkmark$} & \textcolor{lawgreen2}{$\checkmark$} & \textcolor{lawgreen2}{$\checkmark$} & 9 & 9.8\% & \cellcolor{lawgreen2}5.7\% & \cellcolor{lawgreen1}\textbf{0.95} & 2.8\% & 3.7\% & \cellcolor{lawgreen1}0.83 & 1.1\% & \cellcolor{lawgreen2}\textbf{2.1\%} & \cellcolor{lawgreen1}\textbf{0.81} & 4.6\% & \cellcolor{lawgreen2}3.8\% & \cellcolor{lawgreen2}0.86 \\
\bottomrule
\end{tabular}}%
}
\caption{Paired scaling law scores on the three fit tests: percent error for $D\uparrow$ and $p=1$ (lower is better), and within-cell $R^2$ (higher is better).}
\label{tab:law-bench}
\end{table*}

\subsection{Evaluation}
\label{sec:law-eval}

We evaluate each law based on four held-out tests after fitting each equation via least squares to each test's respective training runs. $D_\text{max}$ represents the size of data for the largest run of each environment. The three tests are:

\begin{enumerate}
    \item \textbf{Scale} ($D{\uparrow}$): fit on runs with $D \in [0, D_\text{max}/2)$ and test on runs with $D \in [D_\text{max}/2, D_\text{max}]$
    \item \textbf{Full pairing} ($p=1$): fit on runs where $p \in [0, 1/2)$ and test on the corner $p=1$
    \item \textbf{Pairing shape} ($R^2$): fit on all except one $(D_1, D_2)$ grid, predict every loss for every $p$ at $(D_1, D_2)$, subtract the mean, then measure explained variance
\end{enumerate}

\subsection{Results}
\label{subsec:law-results}

\begin{figure}[t]
\centering
\includegraphics[width=\linewidth]{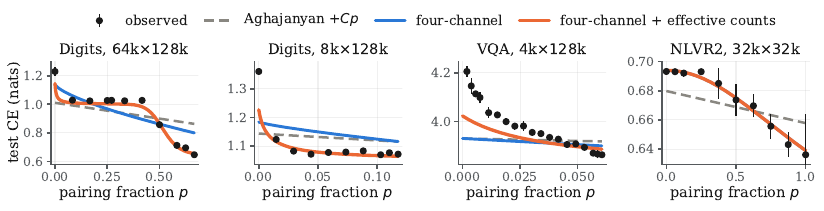}
\caption{Paired scaling laws on four different $D$ grids for the Leave-One-Out test, meaning each curve is fit on other $(D_1, D_2)$ combinations and tested on this $(D_1, D_2)$ cell. All cells, laws, and tests can be viewed \href{https://paired-scaling-laws-viewer.up.railway.app/}{at this web link}.}
\label{fig:law-fits}
\end{figure}

Results of each fit are in Table~\ref{tab:law-bench}. Extensions of published laws are the weakest fits across all tests and environments, with every variant of the four-channel law beating published laws on $D\uparrow$, 14 of 16 on $p=1$, and 13 of 16 on $R^2$. The two factors that are most helpful for the four-channel law are the synergy gate and effective counts. Shared exponents reduce the number of fitted parameters and also improve overall fit. PID loss weights do not improve test performance, although they do reduce the number of fitted parameters.

\paragraph{Theory behind the four-channel law's improvement.} We identify two theoretical reasons why the four-channel law outperforms extensions. Let $\Gamma = -\partial^2 L/\partial D_1 \partial D_2$ at fixed $p$, which represents how much adding an example of modality 1 changes the value of adding an example of modality 2. Properly modeling synergy requires $\Gamma > 0$ at some point, since synergistic information can only be acquired when both modalities are present. Additionally, proper modeling requires $\partial \Gamma / \partial D$ must be non-zero, i.e. $\Gamma$ changes with changing data, as different information channels eventually saturate. Every existing law except the Ye mixture violates either or both of these constraints (Table~\ref{tab:law-candidates}), and Ye only models relative share instead of absolute counts as we observed in \S\ref{sec:paired-curves} and \S\ref{sec:synergy_gates}. The four-channel law contains both properties via the $S$ decomposition, where synergy loss is gated before a threshold (allowing $\Gamma > 0$) and decays (allowing $\partial \Gamma / \partial D \neq 0$).

\paragraph{Interactive viewer.} We display examples of different law fits plotting CE against $p$ in Figure~\ref{fig:law-fits}. Four-channel laws are able to successfully model synergy gates, delayed loss, and $p$-dependent loss. To further illustrate these results, we have included an interactive paired scaling law viewer at \href{https://paired-scaling-laws-viewer.up.railway.app/}{this web link}.

\paragraph{Optimal pairing fractions.} These laws can also be used to determine the best pairing fraction under a budget where pairs are expensive. For NLVR2, which is almost all synergy, optimal pairing is always $p=1$, but for Digits and VQA, it moves well below $p=1$ when pairs cost two to three times unpaired data (Appendix~\ref{app:budget}).

\section{Conclusion}
\label{sec:conclusion}

We investigated how multimodal loss changes as a function of data pairing and found that only pairs can reduce synergistic loss while unpaired data can reduce redundant and modal-unique loss. We introduced a new four-channel multimodal scaling law that outperforms existing scaling laws and also proved that data pairing is an independent variable that has measurable impact on model loss.

\paragraph{Scope and limitations.} Partial information decomposition is defined with respect to a label $Y$, so our laws are most applicable to supervised multimodal tasks. We did not investigate self-supervised generative pretraining without labels where pairing itself is the training signal. We leave investigation of the generalizability of our four-channel scaling law in self-supervised training regimes as future work, as well as larger data and parameter scales. We note many established scaling law papers have been tested at scales comparable to our experiments \citep{hestness_2017, rosenfeld_2020, sorscher_2022, bahri_2024} and that the power-law forms they describe hold at larger scales \citep{kaplan_2020, hoffmann_2022, aghajanyan_2023}.

\clearpage
\subsection*{AI Use Statement}

In this work, we performed the following tasks \textbf{without} the help of AI or LLM usage:

\begin{itemize}
    \item Formulating core mathematical claims and theorems
    \item Developing theoretical models and conceptual frameworks
    \item Proposing and refining hypotheses
    \item Discovering relevant research areas and important existing literature
    \item Designing research methodology and experiments
    \item Interpreting results
    \item Identifying experimental parameters
    \item Verifying validity of claims
    \item Designing the paper structure
    \item Crafting the title and keywords
    \item Writing the final manuscript
\end{itemize}

The following tasks were performed \textbf{with} the help of AI and LLM usage, with brief justification:

\begin{itemize}
    \item Providing critical ingredients and writing of proofs: We used AI and LLMs to double-check and verify the proposed proofs as well as simulate different variations of each equation and check for edge cases.
    \item Providing feedback for research methodology and experiments: After outlining the core experiments, we used AI to help brainstorm potential design flaws and ablations. A human was always present to review and approve ablations.
    \item Implementing methods and creating code: We used AI and LLMs to generate the code to execute our experiments. The majority of our code is AI-written, including the pairing viewer, but the authors have reviewed the code manually and ran unit tests, sanity checks, and manually validated that the execution and results are error-free as best to the authors' ability.
    \item Creating or modifying figures or images: We used AI and LLMs to generate figures and tables, mostly for ease of formatting. All ideas and layouts for figures and tables came from the authors directly. The results and layout were then manually adjusted and refined by the authors.
    \item Cleaning and reformating datasets: We used LLMs to standardize our several datasets in the codebase.
    \item Creating draft sections of the paper: We used AI and LLMs to summarize and create draft sections of several parts of the paper; however, every sentence in the final manuscript is human-written.
    \item Performing related literature sweeps: After the key relevant research areas were identified through manual review, we used AI and LLMs to exhaustively search for all relevant literature.
    \item Grammar and formatting: We used AI and LLMs to perform final grammar and formatting checks in the manuscript.
\end{itemize}

The following tasks are not applicable to this work:

\begin{itemize}
    \item Generating synthetic data sets
    \item Assisting with translation
    \item Supporting qualitative and thematic data analysis
    \item Formulating questions for surveys or interviews
    \item Transcribing recordings
\end{itemize}

We have reviewed all AI-assisted work and take full responsibility for the final content of this work.
\clearpage
\subsection*{Ethics Statement}

We do not believe this work raises any ethical concerns, as this work is solely focused on fundamental machine learning theory research and empirical analysis. This work does not collect new data, deploy a system, or present any potential data privacy concerns. We use existing research datasets in compliance with their usage licenses and do not redistribute external dataset images and text.

\bibliography{references}
\bibliographystyle{iclr2027_conference}

\clearpage

\appendix

\section{Paired scaling-law derivations}
\label{app:paired-laws}

In this appendix section, we derive $\delta$ (\ref{sec:th-bounded}) and $\Gamma$ (\ref{sec:th-gamma}), define effective counts (\ref{sec:app-eff}), show two additional law fit results (\ref{sec:app-law-interp}), and plot optimal pairing ratios under a budget (\ref{app:budget}).

\subsection{Single-Source Loss Curves with $\delta$}
\label{sec:th-bounded}

The Hoffmann form (Eq.~\ref{eq:hoffmann-scaling}) diverges as $D \to 0$ but classification loss is bounded by the loss of an untrained model, $L_0$ ($\ln k$ for $k$ balanced classes). To make this equation well-behaved at low $D$, we add a normalization term $\delta$, which acts as a pseudocount of seen examples:

\begin{equation}
    L(D) = E_N + \frac{B}{(D+\delta)^\beta}.
    \label{eq:th-hoffmann-with-nu}
\end{equation}

At $D = 0$ this gives $L_0 = E_N + B/\delta^\beta$, which rewritten yields $B = (L_0 - E_N)\,\delta^\beta$ and
\begin{equation}
    L(D) = E_N + (L_0 - E_N)\, g(D), \qquad g(D) = \Big(\frac{\delta}{D+\delta}\Big)^{\beta}.
    \label{eq:th-L-refactored}
\end{equation}
For $D \gg \delta$ this approaches the original term $B/D^\beta$. Both endpoints $L_0$ and $E_N$ can be directly measured via expected loss of a uniform distribution and a single long training run that saturates on data and compute, respectively. This leaves the exponents $\delta$ and $\beta$ to be fit instead of $B$ and $\beta$ (as $B$ can derived through $L_0$, $E_N$, $\delta$, and $\beta$). We can thus estimate the value of a single example at a given loss:

\begin{equation}
    \frac{\mathrm{d}L}{\mathrm{d}D} = -\frac{\beta}{D+\delta}\,\big(L(D) - E_N\big).
    \label{eq:th-derivative}
\end{equation}
\citet{jiang_2024} use a similar quantity, without $\delta$, to choose data mixtures during training.

\subsection{Interaction Sign $\Gamma$}
\label{sec:th-gamma}

Let $v_i = -\partial L/\partial D_i$ be the value of adding one more example of modality $i$. This value is always positive as loss curves are monotonically decreasing, and comparing $v_i$ to $v_j$ determines whether modality $i$ or $j$ is more locally helpful. We define an interaction sign $\Gamma$ as
\begin{equation}
    \Gamma = \frac{\partial v_i}{\partial D_j} = -\frac{\partial^2 L}{\partial D_i\, \partial D_j}.
\end{equation}
If $\Gamma > 0$, the modalities are complements: more of $j$ makes an example of $i$ more valuable. If $\Gamma = 0$, they are independent, and if $\Gamma < 0$, they are substitutes. A multimodal law that can express synergy should be able to express all three, as there are scenarios when one modality helps, hurts, or is neutral to additional data from the other modality.

\paragraph{Proof that existing laws (except Ye) have $\Gamma \leq 0$ at fixed $p$.} All Hoffmann-style equations take the form $L = E_N + Wg(D_i + \kappa D_j)$ with $\kappa \geq 0$. $g$ is a decreasing convex curve, thus
\begin{equation}
\frac{\partial^2 L}{\partial D_i \, \partial D_j} = W \kappa\, g''(D_i + \kappa D_j) \ge 0
\qquad\Longleftrightarrow\qquad \Gamma \le 0
\label{eq:th-sign-rigid}
\end{equation}
for every parameter setting. This means Hoffmann-derived equations model modalities $i$ and $j$ as substitutes only. Similarly, at fixed $p$, Aghajanyan's $Cp$ and Shukor's $(C_1 p + C_2(1-p))^{-1}$ are constants, so Aghajanyan's $\Gamma$ is $0$, and Shukor's is negative from its pooled $g(D_{\mathrm{tot}})$. Ye's exponent depends on the share $(D_2 - n_p)/D_{\mathrm{tot}}$, which at fixed $p$ can change because $n_p$ is free even at fixed $p$ by allocating additional $D_1$.

\paragraph{Proof that the four-channel law does not constrain $\Gamma$ at fixed $p$.}

The four-channel family is
\begin{align}
L &= E_N + L_R + L_{U_1} + L_{U_2} + L_S, \\
D_R &= D_1 + \kappa D_2,\quad
D_{U_1} = D_1,\quad
D_{U_2} = D_2,\quad
D_S = n_p.
\end{align}
With power-law channels, $L_a = B_a g(D_a)$, every term is a convex function of one count or of a sum. The base law therefore has $\Gamma \le 0$ like the Hoffmann forms. However, adding the synergy gate replaces $L_S$ with $B_S/(1+(n_p/n_\star)^\gamma)$. For $\gamma > 1$ this curve is concave below $n_p = n_\star\big((\gamma-1)/(\gamma+1)\big)^{1/\gamma}$, so there $L_S$ contributes $-(p/2)^2 L_S'' > 0$ to $\Gamma$. Before synergy is learned, more of either modality makes the other more valuable, because together they buy the pairs that open the gate. After the gate, the curve is convex again and the $\Gamma$ becomes negative. Effective counts change the counts that feed each channel and can in rare cases make $\Gamma$ positive.

\subsection{Effective counts}
\label{sec:app-eff}

A model trained without pairs can model unimodal channels effectively but see increased loss during test-time paired data (\S\ref{subsec:presence-matching}). We model this as unpaired data being less valuable until a certain amount of paired data is available to align paired data with unpaired representations. At zero pairs, an unpaired example is worth $\lambda_0 \le 1$ of a paired one and effective count is represented as:
\begin{equation}
    D_i^{\mathrm{eff}} = D_i - (1-\lambda_0)\, n_i\, \frac{n_\lambda}{n_p + n_\lambda}.
    \label{eq:eff-counts}
\end{equation}
$n_i = D_i - n_p$ is the unpaired count of modality $i$ and $n_\lambda$ is the number of pairs that unlocks half the locked value. More pairs increases the effective counts of unpaired data, and thus pairs and unpaired examples become complements initially.

\subsection{Additional Law Fit Tests}
\label{sec:app-law-interp}

In Table~\ref{tab:law-interp}, we report two interpolation law fits: the Leave-One-Out test, where an equation curve is fit on all cells except one ($D_1$, $D_2$) grid and then tested on every $p$ of that heldout ($D_1$, $D_2$) cell, and the Full test, where no cells are heldout and one law is fit over every data point. These two tests reveal the same results as the main paper, that the four-channel law outperforms extensions of existing laws.

\begin{table*}[t]
\centering
\scriptsize
\definecolor{lawred2}{RGB}{198,96,96}
\definecolor{lawred1}{RGB}{246,210,210}
\definecolor{lawgreen1}{RGB}{208,234,212}
\definecolor{lawgreen2}{RGB}{96,168,112}
{
\setlength{\aboverulesep}{0pt}
\setlength{\belowrulesep}{0pt}
\renewcommand{\arraystretch}{1.0}
\resizebox{\textwidth}{!}{%
\begin{tabular}{@{}c *{4}{c} c *{8}{wc{0.95cm}}@{}}
\toprule
& \multicolumn{4}{c}{} & & \multicolumn{2}{c}{Digits} & \multicolumn{2}{c}{VQA} & \multicolumn{2}{c}{NLVR2} & \multicolumn{2}{c}{Overall} \\
\cmidrule(lr){7-8}\cmidrule(lr){9-10}\cmidrule(lr){11-12}\cmidrule(lr){13-14}
& \multicolumn{4}{c}{Law} & $k$ & LOO & Full & LOO & Full & LOO & Full & LOO & Full \\
\midrule
 & \multicolumn{4}{l}{Hoffmann Unimodal} & 4 & \cellcolor{lawred2}5.9\% & \cellcolor{lawred2}5.6\% & \cellcolor{lawred2}3.9\% & \cellcolor{lawred2}3.7\% & \cellcolor{lawred2}1.0\% & \cellcolor{lawred2}1.3\% & \cellcolor{lawred2}3.6\% & \cellcolor{lawred2}3.5\% \\
 & \multicolumn{4}{l}{Hoffmann on unique $D$} & 4 & \cellcolor{lawred2}5.2\% & \cellcolor{lawred1}4.5\% & \cellcolor{lawred2}4.0\% & \cellcolor{lawred2}3.8\% & \cellcolor{lawred2}1.0\% & \cellcolor{lawred2}1.2\% & \cellcolor{lawred2}3.4\% & \cellcolor{lawred2}3.1\% \\
 & \multicolumn{4}{l}{Hoffmann on $D_1$ and $D_2$} & 7 & 4.2\% & 3.6\% & 1.1\% & 1.0\% & \cellcolor{lawred2}1.2\% & \cellcolor{lawred2}1.3\% & 2.2\% & 2.0\% \\
 & \multicolumn{4}{l}{Aghajanyan $+Cp$} & 8 & \cellcolor{lawred2}5.7\% & \cellcolor{lawred1}4.6\% & 1.2\% & 1.0\% & \cellcolor{lawred2}0.9\% & \cellcolor{lawred1}0.8\% & \cellcolor{lawred1}2.6\% & \cellcolor{lawred1}2.2\% \\
 & \multicolumn{4}{l}{Shukor mixture} & 6 & 4.1\% & 3.9\% & \cellcolor{lawred2}4.6\% & \cellcolor{lawred2}4.3\% & \cellcolor{lawred1}0.8\% & \cellcolor{lawred2}1.1\% & \cellcolor{lawred2}3.2\% & \cellcolor{lawred2}3.1\% \\
 & \multicolumn{4}{l}{Ye mixture} & 7 & \cellcolor{lawred1}4.4\% & 4.1\% & \cellcolor{lawred1}2.0\% & 1.7\% & \cellcolor{lawred2}0.9\% & \cellcolor{lawred2}1.0\% & \cellcolor{lawred1}2.4\% & \cellcolor{lawred1}2.3\% \\
\midrule
\multirow{17}{*}{\rotatebox[origin=c]{90}{\small Four-channel}} & Shared & Gate & PID & Eff. & \multicolumn{9}{c}{} \\
\cmidrule(lr){2-2}\cmidrule(lr){3-3}\cmidrule(lr){4-4}\cmidrule(lr){5-5}
 & \textcolor{lawred2}{$\times$} & \textcolor{lawred2}{$\times$} & \textcolor{lawred2}{$\times$} & \textcolor{lawred2}{$\times$} & 14 & \cellcolor{lawred1}4.5\% & \cellcolor{lawred1}4.2\% & 0.8\% & \cellcolor{lawgreen1}\textbf{0.6\%} & 0.5\% & 0.5\% & 1.9\% & 1.8\% \\
 & \textcolor{lawred2}{$\times$} & \textcolor{lawred2}{$\times$} & \textcolor{lawred2}{$\times$} & \textcolor{lawgreen2}{$\checkmark$} & 16 & \cellcolor{lawgreen1}2.8\% & \cellcolor{lawgreen1}2.6\% & \cellcolor{lawgreen1}\textbf{0.7\%} & \cellcolor{lawgreen1}\textbf{0.6\%} & 0.5\% & 0.5\% & \cellcolor{lawgreen1}1.4\% & \cellcolor{lawgreen1}1.2\% \\
 & \textcolor{lawred2}{$\times$} & \textcolor{lawgreen2}{$\checkmark$} & \textcolor{lawred2}{$\times$} & \textcolor{lawred2}{$\times$} & 14 & \cellcolor{lawgreen1}3.2\% & \cellcolor{lawred1}4.2\% & 0.8\% & 0.7\% & \cellcolor{lawgreen2}\textbf{0.2\%} & \cellcolor{lawgreen1}\textbf{0.3\%} & \cellcolor{lawgreen1}1.4\% & 1.7\% \\
 & \textcolor{lawred2}{$\times$} & \textcolor{lawgreen2}{$\checkmark$} & \textcolor{lawred2}{$\times$} & \textcolor{lawgreen2}{$\checkmark$} & 16 & \cellcolor{lawgreen2}\textbf{1.7\%} & \cellcolor{lawgreen2}\textbf{1.4\%} & 0.8\% & \cellcolor{lawgreen1}\textbf{0.6\%} & \cellcolor{lawgreen2}\textbf{0.2\%} & \cellcolor{lawgreen1}\textbf{0.3\%} & \cellcolor{lawgreen2}0.9\% & \cellcolor{lawgreen2}\textbf{0.8\%} \\
 & \textcolor{lawgreen2}{$\checkmark$} & \textcolor{lawred2}{$\times$} & \textcolor{lawred2}{$\times$} & \textcolor{lawred2}{$\times$} & 8 & \cellcolor{lawred2}5.2\% & \cellcolor{lawred2}4.9\% & \cellcolor{lawgreen1}0.8\% & 0.7\% & 0.5\% & 0.5\% & 2.1\% & 2.0\% \\
 & \textcolor{lawgreen2}{$\checkmark$} & \textcolor{lawred2}{$\times$} & \textcolor{lawred2}{$\times$} & \textcolor{lawgreen2}{$\checkmark$} & 10 & 4.1\% & 3.7\% & \cellcolor{lawgreen1}\textbf{0.7\%} & \cellcolor{lawgreen1}\textbf{0.6\%} & 0.5\% & 0.5\% & 1.8\% & 1.6\% \\
 & \textcolor{lawgreen2}{$\checkmark$} & \textcolor{lawgreen2}{$\checkmark$} & \textcolor{lawred2}{$\times$} & \textcolor{lawred2}{$\times$} & 10 & 3.3\% & 3.9\% & 0.8\% & 0.7\% & \cellcolor{lawgreen2}\textbf{0.2\%} & \cellcolor{lawgreen1}\textbf{0.3\%} & \cellcolor{lawgreen1}1.4\% & 1.6\% \\
 & \textcolor{lawgreen2}{$\checkmark$} & \textcolor{lawgreen2}{$\checkmark$} & \textcolor{lawred2}{$\times$} & \textcolor{lawgreen2}{$\checkmark$} & 12 & \cellcolor{lawgreen2}\textbf{1.7\%} & \cellcolor{lawgreen2}\textbf{1.4\%} & \cellcolor{lawgreen1}\textbf{0.7\%} & \cellcolor{lawgreen1}\textbf{0.6\%} & \cellcolor{lawgreen2}\textbf{0.2\%} & \cellcolor{lawgreen1}\textbf{0.3\%} & \cellcolor{lawgreen2}\textbf{0.8\%} & \cellcolor{lawgreen2}\textbf{0.8\%} \\
 & \textcolor{lawred2}{$\times$} & \textcolor{lawred2}{$\times$} & \textcolor{lawgreen2}{$\checkmark$} & \textcolor{lawred2}{$\times$} & 11 & \cellcolor{lawred1}4.6\% & \cellcolor{lawred1}4.3\% & 0.8\% & \cellcolor{lawgreen1}\textbf{0.6\%} & 0.5\% & 0.5\% & 2.0\% & 1.8\% \\
 & \textcolor{lawred2}{$\times$} & \textcolor{lawred2}{$\times$} & \textcolor{lawgreen2}{$\checkmark$} & \textcolor{lawgreen2}{$\checkmark$} & 13 & 3.5\% & 3.1\% & \cellcolor{lawgreen1}\textbf{0.7\%} & 0.7\% & 0.5\% & 0.5\% & 1.6\% & 1.4\% \\
 & \textcolor{lawred2}{$\times$} & \textcolor{lawgreen2}{$\checkmark$} & \textcolor{lawgreen2}{$\checkmark$} & \textcolor{lawred2}{$\times$} & 11 & 3.3\% & \cellcolor{lawred1}4.3\% & 0.8\% & 0.7\% & \cellcolor{lawgreen2}\textbf{0.2\%} & \cellcolor{lawgreen1}\textbf{0.3\%} & \cellcolor{lawgreen1}1.4\% & 1.8\% \\
 & \textcolor{lawred2}{$\times$} & \textcolor{lawgreen2}{$\checkmark$} & \textcolor{lawgreen2}{$\checkmark$} & \textcolor{lawgreen2}{$\checkmark$} & 13 & \cellcolor{lawgreen1}2.7\% & \cellcolor{lawgreen2}1.6\% & \cellcolor{lawgreen1}\textbf{0.7\%} & \cellcolor{lawgreen1}\textbf{0.6\%} & \cellcolor{lawgreen2}\textbf{0.2\%} & \cellcolor{lawgreen1}\textbf{0.3\%} & \cellcolor{lawgreen1}1.2\% & \cellcolor{lawgreen2}\textbf{0.8\%} \\
 & \textcolor{lawgreen2}{$\checkmark$} & \textcolor{lawred2}{$\times$} & \textcolor{lawgreen2}{$\checkmark$} & \textcolor{lawred2}{$\times$} & 5 & 3.4\% & \cellcolor{lawgreen1}2.9\% & 1.4\% & 1.2\% & 0.5\% & 0.5\% & 1.7\% & 1.5\% \\
 & \textcolor{lawgreen2}{$\checkmark$} & \textcolor{lawred2}{$\times$} & \textcolor{lawgreen2}{$\checkmark$} & \textcolor{lawgreen2}{$\checkmark$} & 7 & \cellcolor{lawred1}4.8\% & \cellcolor{lawred1}4.7\% & 0.9\% & 0.8\% & 0.5\% & 0.5\% & 2.1\% & 2.0\% \\
 & \textcolor{lawgreen2}{$\checkmark$} & \textcolor{lawgreen2}{$\checkmark$} & \textcolor{lawgreen2}{$\checkmark$} & \textcolor{lawred2}{$\times$} & 7 & 3.5\% & \cellcolor{lawgreen1}2.8\% & \cellcolor{lawgreen1}\textbf{0.7\%} & 0.7\% & \cellcolor{lawgreen2}\textbf{0.2\%} & \cellcolor{lawgreen1}\textbf{0.3\%} & \cellcolor{lawgreen1}1.5\% & \cellcolor{lawgreen1}1.3\% \\
 & \textcolor{lawgreen2}{$\checkmark$} & \textcolor{lawgreen2}{$\checkmark$} & \textcolor{lawgreen2}{$\checkmark$} & \textcolor{lawgreen2}{$\checkmark$} & 9 & \cellcolor{lawgreen2}2.3\% & \cellcolor{lawgreen2}1.9\% & \cellcolor{lawgreen1}\textbf{0.7\%} & 0.7\% & \cellcolor{lawgreen2}\textbf{0.2\%} & \cellcolor{lawgreen1}\textbf{0.3\%} & \cellcolor{lawgreen2}1.1\% & \cellcolor{lawgreen2}0.9\% \\
\bottomrule
\end{tabular}}%
}
\caption{Paired scaling law scores of the two interpolation tests: Leave-One-Out and the Full test.}
\label{tab:law-interp}
\end{table*}

\subsection{Choosing a Pairing Fraction Under a Budget}
\label{app:budget}

\begin{figure}[t]
\centering
\includegraphics[width=\linewidth]{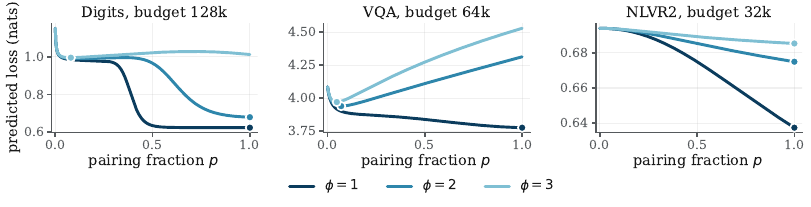}
\caption{Predicted loss versus pairing fraction $p$ at pair prices $\phi \in \{1, 2, 3\}$. Dots mark optima.}
\label{fig:pairing-budget}
\end{figure}

We consider the practical problem of the optimal pairing ratio under a resource budget where pairs cost more than unpaired data. Figure~\ref{fig:pairing-budget} plots loss versus $p$ given a pair price $\phi \in \{1, 2, 3\}$ for the three environments. When $\phi=1$ (i.e. pairs do not cost extra), optimal $p=1$, but higher $\phi$ changes $p$ depending on the environment: for NLVR2, which is dominated by synergy, optimal $p$ is always 1, but for Digits and VQA, this value lies in the interior.

\section{Additional Experiment Implementation Details}
\label{app:experiment-implementation}

In the sections below, we provide additional implementation detail for each of the environments.

\subsection{Digits Experiment}

\paragraph{Labels and known constants.}
Each modality shows four digits, one per label position (\S\ref{sec:experiments}). The image is four MNIST digits side by side ($1 \times 28 \times 112$). The audio is four AudioMNIST log-mel spectrograms joined in time ($1 \times 64 \times 400$). Training audio comes from speakers 1--48 and test audio from speakers 49--60. Test images come from the MNIST test split. Symmetric label noise at rate $\varepsilon = 0.1$ is applied to the label only. Output heads are initialized to zero so that an untrained model predicts the uniform distribution.

\paragraph{Model.}
For each modality, we use the same encoder: three blocks of a $3\times3$ convolution (32, 64, and 128 channels and max-pooling), then GroupNorm and ReLU. The encoder pools to a single column per digit position and then projects to a 128-d embedding. We train with auxiliary cross-entropy of weight $1$ for modality presence to aid with modular sum for the synergy digit. Missing modalities are replaced with a null embedding. The fusion trunk is a two-layer neural network with hidden layer width of 256, which sees both embeddings and the modality presence bits, then feeds into four ten-way heads, one for each label digit. This model has $483k$ trainable parameters.

\paragraph{Training and evaluation.} We used AdamW with learning rate $3\times10^{-4}$ and weight decay $10^{-2}$, batch size 128 drawn uniformly with replacement, cosine decay over the planned number of steps, and gradient clipping at $2$. We hold out one tenth of training for a validation set, used for logging. We trained for $k \in \{10, 20, 40\}$ epochs. The test set is $5000$ examples.

\subsection{VQA-v2 Experiment}

\paragraph{Data.}
The train split of VQA-v2 has $443,757$ questions on COCO train204 images with 10 annotator answers per question \citep{antol_2015,goyal_2017}. The vocabulary is the $3{,}129$ answers that occur at least 64 times. The $8{,}872$ questions with no answer in the vocabulary are omitted, leaving $434{,}885$ questions on $82{,}772$ images. The target label is the soft label of annotator votes, and loss is the cross-entropy in nats against the soft label. The test set is  $25{,}000$ val2014 questions. 

\paragraph{Unpaired examples.} We define an unpaired image as an image with no text, with the answer label sampled independently from any of the questions that are asociated with that image. An unpaired text example is a normal question example with the image nulled.

\paragraph{Model.} We used CLIP ViT-B/32 \citep{radford2021clip} to embed images and questions one time into a frozen cache. Using these embeddings, we trained an MLP head ($1026 \to w \to w \to 3129$ then GELU) over the two embeddings and two presence bits. In the main experiment, $w = 1024$ ($5.3$M parameters), and in Appendix~\ref{app:architecture} we test $w=512$ and $256$ ($2.39$M and $1.13$M). The last layer is zero-initialized. We used AdamW, learning rate $3\times10^{-4}$, weight decay $0.01$, batch size 512. One sixteenth of training data was heldout as validation.

\subsection{NLVR2 Experiment}

\paragraph{Data.}
We used the LXMERT release of NLVR2 train, which contains $86{,}373$ examples \citep{suhr_2019,tan_2019}. Each example is two photographs, one sentence, and a true/false label. The test set is the official development split of $6{,}982$ examples. A photos-only example has both photographs, no sentence, and the label. A sentence-only example has the sentence, no photographs, and the label.

\paragraph{Model.}
We used LXMERT-base \citep{tan_2019} which has $210$M parameters. Finetuning backpropogated through the entire model. After encoding both photographs, the two outputs were concatenated into a two-layer classifier. We used BertAdam optimizer with learning rate $5\times10^{-5}$, 10\% warmup and linear decay, weight decay $0.01$, gradient clipping at 5, batch size 32, sentence length 20, label smoothing $0.1$, and a fresh shuffle each epoch. One sixteenth of training data was heldout as validation.

\paragraph{Metric.}
We used test cross-entropy as our loss after fitting temperature on validation logits, as we encountered improved accuracy but worsened cross-entropy in initial runs due to overconfidence.

\section{Robustness Checks}
\label{app:robustness}

We test several factors for robustness of our claims in the main paper. We report accuracy instead of cross-entropy (\ref{app:metrics}), vary architecture and model size (\ref{app:architecture}), and change how we count paired data (\ref{app:counting}).

\subsection{Accuracy Instead of Cross-Entropy}
\label{app:metrics}

\begin{figure}[t]
\centering
\includegraphics[width=\linewidth]{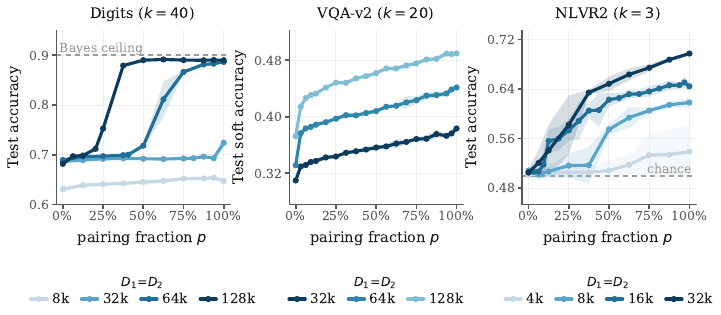}
\caption{Figure~\ref{fig:pair-curves} in accuracy instead of test cross-entropy.}
\label{fig:acc-curves}
\end{figure}

\begin{figure}[t]
\centering
\includegraphics[width=\linewidth]{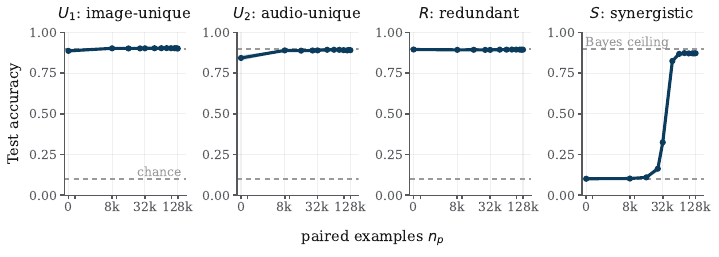}
\caption{Figure~\ref{fig:pair-slots} (Top) in accuracy instead of cross-entropy. Per-position test accuracy in Digits at $D_1 = D_2 = 128$k and $k = 40$ with both modalities shown.}
\label{fig:acc-slots}
\end{figure}

Figure~\ref{fig:acc-curves} repeats Figure~\ref{fig:pair-curves} in accuracy instead of cross-entropy, and Figure~\ref{fig:acc-slots} repeats Figure~\ref{fig:pair-slots} in per-position accuracy in Digits. Both metrics yield similar findings.
\begin{itemize}
    \item \textbf{Synergy is gated.} In Figure~\ref{fig:acc-curves}, we observe delayed improved test accuracy for Digits and NLVR2, with no such gate for VQA excepting the $p=0$ jump. This is the same as in cross-entropy.
    \item \textbf{VQA displays a $p=0$ jump.} In VQA, accuracy rises from $0.37$ to $0.42$ with the first $4$k pairs, then rises steadily to $0.49$ at $p = 1$.
    \item \textbf{Presence-matching has a smaller effect for accuracy.} Unlike test cross-entropy, pairing has minimal effect on unimodal and redundant accuracy (Figure~\ref{fig:acc-slots}); at this data budget $D_1=D_2=128$k, the $U_1$, $U_2$, and $R$ slots are accurate up to the label noise threshold of $90\%$.
\end{itemize}

\subsection{Architecture and Model Size}
\label{app:architecture}

\paragraph{Digits.} We replace the CNN encoders from Digits with two other encoders of different architectures, both trained from scratch on the same data: an MLP and a transformer. The MLP contains two layers of width 256 for each input digit and the transformer has sixteen patches per digit position with width 128, four layers, and four heads.

\begin{figure}[t]
\centering
\includegraphics[width=0.72\linewidth]{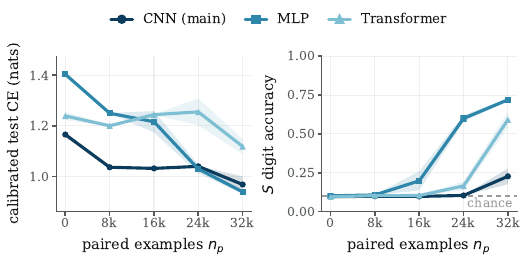}
\caption{Digits CE (left) and $S$ digit accuracy (right) versus $n_p$ with three different encoders at $D_1 = D_2 = 32$k and $k = 40$.}
\label{fig:arch-digits}
\end{figure}

Figure~\ref{fig:arch-digits} shows loss curves at $D_1 = D_2 = 32$k for the MLP and transformer architecture that are consistent with the CNN encoder. Test CE drops before synergy is learned, and $S$ digit accuracy stays at chance then rises after a threshold. We note that architecture does seem to change where the threshold $n_p$ is, with $S$ digit accuracy first leaving chance at $16$k pairs for the MLP, at $24$k pairs for the transformer, and at $32$k for the CNN.

\begin{figure}[t]
\centering
\includegraphics[width=\linewidth]{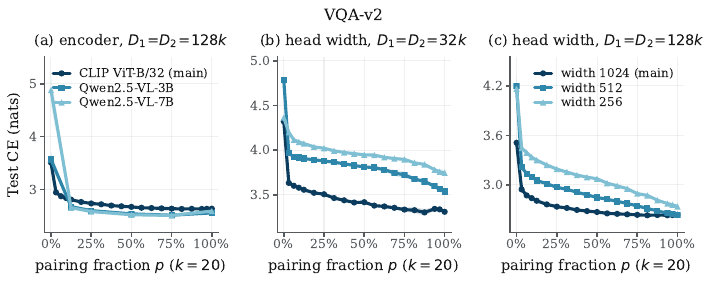}
\caption{VQA-v2 test cross-entropy against $p$ at $k = 20$. \textbf{(a)} Three frozen encoders under the width-1024 head at $D_1 = D_2 = 128$k. \textbf{(b, c)} Three head widths on CLIP features at $32$k and $128$k.}
\label{fig:arch-vqa}
\end{figure}

\paragraph{VQA-v2 encoders.} We replace frozen CLIP features with frozen Qwen2.5-VL-3B and Qwen2.5-VL-7B features (2048 and 3584 dimensions) under the same width-1024 head, at $D_1 = D_2 = 128$k. The Qwen image vector is the mean-pooled output of the vision encoder. The Qwen text vector is the mean of the question's input token embeddings with no transformer layers applied. Figure~\ref{fig:arch-vqa} (a) shows the same shape for all three encoders of a steep drop after $p=0$ then a steady and gradual fall until $p=1$.

\paragraph{VQA-v2 head width.} Figures~\ref{fig:arch-vqa} (b) and (c) vary the MLP head width to 512 and 256. Each width observes the same behavior, and at fully paired data at $D_1=D_2=128$k, all three heads end within $0.1$ nats of each other. Model width changes the initial drop and gradual decline of the curve but not the curve shape.

\subsection{Counting paired and unpaired data}
\label{app:counting}

The main text counts data per modality as $D_1 = n_1 + n_p$ and $D_2 = n_2 + n_p$. In this regime, a paired datum counts towards both $D_1$ and $D_2$. There are two other ways to count data:
\begin{itemize}
    \item \textbf{Counting by label}, $D_1 + D_2 - n_p = n_1 + n_2 + n_p$, where a paired datum counts once.
    \item \textbf{Counting by price}, where a pair costs $\phi$ times as much as one unpaired example of each modality (Appendix~\ref{app:budget}). If measuring by label and using the original definitions of $D_1$ and $D_2$, then a pair is actually cheaper than an unpaired datum with $\phi = \tfrac12$.
\end{itemize}

\begin{figure}[t]
\centering
\includegraphics[width=\linewidth]{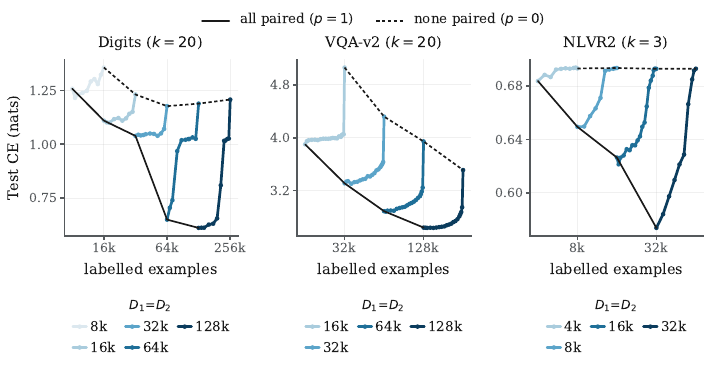}
\caption{The runs of Figure~\ref{fig:pair-curves} replacing the $x$-axis with number of labelled examples, $n_1 + n_2 + n_p = 2D - n_p$, instead of pairing fraction. Each colored curve represents one matched budget $D_1 = D_2 = D$ running from $p = 0$ (right point, $2D$ labeled examples) to $p = 1$ (left point, $D$ labeled examples).}
\label{fig:count-rows}
\end{figure}

\paragraph{Consequences of counting by label.} Because the laws take $(D_1, D_2, n_p)$ as input, the above decision of counting by label does not change any of the law fits, but it does change what ``equal in size'' means as well as deciding an optimal $p$ under a budget. In Figure~\ref{fig:count-rows}, we replot Figure~\ref{fig:pair-curves} with number of labeled examples as the x-axis instead of paired examples. We find in this case, pairing is even more beneficial, as in addition to the synergistic gains demonstrated in this work, paired data also reduces the number of overall labels. This way of counting makes pairs cheaper and biases the optimal pairing ratio under a fixed budget towards more pairs.

\end{document}